\pdfoutput=1
\documentclass[11pt]{article}

\usepackage[preprint]{acl}

\usepackage{times}
\usepackage{latexsym}

\usepackage[T1]{fontenc}
\usepackage[utf8]{inputenc}

\usepackage{microtype}

\usepackage{inconsolata}

\usepackage{graphicx}
\usepackage{booktabs}
\usepackage{multirow}
\usepackage{multicol}
\usepackage{amsfonts}
 \usepackage{amsmath}
 \usepackage{pifont}
\title{Spatial-Aware Efficient Projector for MLLMs via Multi-Layer Feature Aggregation}

\author{ Shun Qian\textsuperscript{1}, Bingquan Liu\textsuperscript{1}, Chengjie Sun\textsuperscript{1}\\ \textbf{Zhen Xu\textsuperscript{1}, Baoxun Wang\textsuperscript{2}} \\
\textsuperscript{1}Harbin Institute of Technology, Harbin, China \\
\textsuperscript{2}Tencent PCG \\
  }

\begin{document}
\maketitle
\begin{abstract}
%%% projector在mllm中承担非常重要的作用，其输出的visual token的数量会影响mllm的效率，visual token的质量会影响llm对图像信息的理解；
%% 尽管目前已有一些工作探索减少visual token的数量来提高mllm的效率，但是通常是以性能为代价。
%% 目前的一些关于projector的探索工作只专注于减少visual token的数量以提高效率，忽略了当前mllm框架下，输入到llm中的序列化的2维视觉信息和自然文本序列之间存在的天然空间关系不一致问题，我们提出了一个能够spatial-aware efficient projector.
%% 使得mllm在大幅降低visual token数量的前提下，在空间关系上取得巨大改善，同时在通用多模态能力评估上，和其他projector相比也取得sota的性能表现。
The projector plays a crucial role in multi-modal language models (MLLMs). The number of visual tokens it outputs affects the efficiency of the MLLM, while the quality of the visual tokens influences the visual understanding capabilities of the MLLM.
Current explorations on the projector focus on reducing the number of visual tokens to improve efficiency, often overlooking the inherent spatial discrepancy between the serialized 2-dimensional visual token sequences and natural language token sequences. 
A \textbf{S}patial-\textbf{A}ware \textbf{E}fficient \textbf{P}rojector (SAEP) is proposed to address this issue.
In detail, our SAEP method employs an modified separable depthwise convolution module on multi-layer visual features to enhance the spatial information of visual tokens.
As a result, our SAEP method can not only largely reduce the number of visual tokens by 75\%, but also significantly improve the multimodal spatial understanding capability of MLLMs.
Moreover, 
compared to existing projectors,
our SAEP gets best performances on massive multimodal evaluation benchmarks, which denotes its effectiveness on bridging the modality gap.
%the strong performance of our SAEP on massive multimodal evaluation benchmarks denotes its effectiveness on bridging the modality gap.

\end{abstract}

\section{Introduction}

Large Language Models (LLMs)~\citep{chenpali, achiam2023gpt, touvron2023llama2, vicuna2023, yang2024qwen2,jiang2024mixtral}have demonstrated marvelous language understanding and reasoning capabilities, making a great impact on the entire AI community.
Meanwhile, many vision foundation models~\citep{radford2021clip, oquab_dinov2, zhai2023siglip, bai2024sequential} aim to produce general-purpose visual features, which work out of the box on any vision task.
As a result,
multimodal works~\citep{liu2023llava, li2023blip2, liu2024llava-next, li2024monkey, dong2024internlm-xcomposer2} introduce a projector to map the visual features into the token embedding space of the LLMs so that the pre-trained knowledge and abilities of both LLMs and vision models can be leveraged.
Multimodal Large Language Models (MLLMs) developed by such a simple technique show powerful visual instruction-following capabilities and largely raise the performance bar of various vision-language tasks.

\begin{table*}[t]
\centering
\small
\begin{tabular}{l|c|llll}
\toprule
Method & \#Token & \textbf{S-Avg.} & \textbf{Ref.-Avg}  & \textbf{VQAv2}  & \textbf{MMB} \\
\midrule
MLP & 576 & 46.7 & 51.7 & 78.5 & 64.3  \\
\midrule
SoTA & 144 & 46.8~\citep{li2024tokenpacker} & 57.6~\cite{chu2024ldpv2} & 78.1~\citep{li2024tokenpacker} & \textbf{66.4}~\citep{chu2024ldpv2} \\
SAEP (Ours)  &  144 & \textbf{51.4} & \textbf{61.4} & \textbf{78.5} & 66.3 \\
\bottomrule
\end{tabular}
\caption{Results of the MLP projector, state-of-the-art efficient projector and ours. They are pre-trained with the same LLavA-1.5 training recipe for a fair comparison. ``S-Avg.'' refers to the normalized average score on 7 multimodal spatial tasks. ``Ref.-Avg'' refers to the average score on the visual grounding task over the RefCOCO/+/g datasets. 
}
\label{tab: first_table}
\end{table*}

Apparently, that the projector plays a crucial role in bridging the modality gap.
It needs to convert the visual features into an ordered visual token sequence so that the visual features can be understood by LLMs~\citep{cha2024honeybee}.
The number of visual tokens directly affects the computational efficiency of the MLLMs, while the quality of visual tokens affects the visual understanding capability of MLLMs.
Despite the importance of the projector, there are few in-depth explorations of it.

As one of the most popular projectors, the two-layer MLP outputs the visual token sequence through the linear transformation.
The MLP projector performs well in MLLMs~\citep{liu2023llava, liu2024llava-next,  liu2023llava1.5} because the one-to-one transformation excels at preserving all local contexts of visual features.
%However, the MLP projector is unable to control the number of visual tokens.
Nonetheless, there are two drawbacks in MLP projector that cannot be ignored.
The first one is that MLP projector cannot control the number of visual tokens, 
thereby affecting the computational efficiency of MLLMs.
This drawback has become much more prominent recently because a group of works~\citep{liu2024llava-next, dong2024internlm2-4khd, li2024monkey, li2023otterhd} has found that the MLLMs can benefit from scaling up the resolution of input images.
Unfortunately, if the resolution of the input image is doubled, there will be a sixteen-fold increase in time complexity and an inevitable explosion in memory requirements in MLLMs.
The second one is that the MLP projector cannot capture spatial knowledge implicated in visual features for LLMs.
In detail,
the visual encoder~\citep{radford2021clip, dosovitskiy2020vit, oquab_dinov2} splits and serializes the input image into a group of patch-level visual features in raster order, which is different from the left-to-right order of natural language.
The simple linear transformation of MLP projector cannot alleviate this spatial and positional discrepancy and limits the performance of MLLMs.
Some projectors~\cite{jaegle2021resampler, li2023blip2, cha2024honeybee, chu2024ldpv2, li2024tokenpacker} are proposed to reduce the number of visual tokens, while the second problem is usually ignored.
As a result, even though these works can reduce the vision token number by 75\% $\sim$ 90\%, 
their performance is limited and is usually worse than the MLP projector.

%%% 已有方案是忽视了空间任务，才限制了其效果的，我们在空间任务上取得非常明显的提升，对通用任务的表现也有帮助. 这样来体现空间任务的重要性

% Some efficient projectors~\citep{cha2024honeybee} focus on reducing the number of visual tokens, while the second problem are usually ignored.
% Therefore, 
% %they reduce the number of vision tokens at the expense of performance.
% even though these works can reduce the vision token number by 75\% $\sim$ 90\%,
% their performance is limited and is usually worse than the MLP projector.

Therefore,  
we propose a \textbf{S}patial-\textbf{A}ware \textbf{E}fficient \textbf{P}rojector, relying on multi-level features, named SAEP.
It can not only largely reduce the number of visual token but also significantly improve the visual spatial understanding capabilities of MLLMs, as shown in Table~\ref{tab: first_table}.
The performance on the general multimodal benchmarks are also improved.
To be specific,
the SAEP projector first reorganizes the patch-level visual feature sequence into a group of  2D ``feature maps'' according to their original positions.
Multi-level visual features are also introduced to provide more detailed local fine grained visual cues and enhance the feature diversity without extra computational overhead.
A modified depthwise separable convolution operation is then applied on these feature maps to compress the local multiple features into a compact one, thereby enabling the preservation of spatial knowledge.

Massive multimodal evaluation benchmarks are utilized to investigate the effectiveness of our SAEP projector from different aspects.
The results indicate that our SAEP projector can not only improve the computational efficiency by reduce 75\% $ \sim $ 89\% visual tokens,
but also demonstrate extraordinary spatial understanding capability.
Moreover, our SAEP projector achieves state-of-the-art performance against existing efficient projectors on the general multimodal benchmarks.

\section{Related Works}
%%% github上有一些压缩token的工作 
\subsection{Multimodal Large Language Models}
Inspired by the great language understanding capabilities exhibited by Large Language Models (LLMs),
the multimodal researchers take pre-trained LLMs as the pivot for information understanding and reasoning, and empower LLMs with multimodal perception capabilities to ``feel'' the  real world and perform various multimodal tasks~\citep{li2023blip2, liu2023llava, liu2023llava1.5}.
The pioneer works such as Flamingo~\citep{alayrac2022flamingo} and BLIP2~\citep{li2023blip2}, focus on bridge the modality gap between the pre-trained unimodal models with massive image-text pairs.
LLaVA~\citep{liu2023llava} first attempts to endow the MLLMs with the ability to visual instruction-following by tuning the model with multimodal instruction-following data samples.
%% 高分辨率图像 %% 交叉图文能力
Many subsequent works improve the MLLMs from different perspectives.
Monkey~\citep{li2024monkey}, OtterHD~\citep{li2023otterhd}, LLaVA-NeXT~\citep{liu2024llava-next}, InternLM-XComposer2-4KHD~\citep{dong2024internlm2-4khd} take high-resolution images as visual input and get noticeable performance improvement.
InternLM-XComposer series~\citep{zhang2023internlm-xcomposer, dong2024internlm-xcomposer2, dong2024internlm2-4khd}, LLaVA-NeXT-Interleave~\citep{li2024llava-next-interleave} and Chameleon~\citep{team2024chameleon} focus on generate interleaved image-text contents.
Mini-Genimi~\citep{li2024mini-genimi}, SPHINX-X~\citep{gao2024sphinx-x}, Prismatic-VLM~\citep{karamcheti2024prismatic} employs multiple different vision encoders to capture diverse visual representations.

\subsection{Vision-Language Projector}
%% 可以将现有token压缩方案分成两组：attention， pooling
%% resampler, q-former, c-abstractor, tokenpacker, LDP, LDPv2
%% llm和visual encoder都用了大量数据做训练，用一个contector将两者结合起来就能以较低成本利用模型已经学到的知识，
%% 这也是目前mllm领域的主流思路，（利用clip是主流要提及）
%% blip2最先提出q-former，Flamingo利用resampler提取特征，设计了一套gated fusion方案
%% llava的工作则显示只需要简单的mlp就能取得明显的效果
%% 考虑到增加图像分辨率能有效提升模型表现，同时带来巨大的计算资源消耗
%To utilize the learned knowledge of pre-trained visual encoder and LLMs, the connector module is 
%% to bridge the visual embedding space to the language space
To align the visual representation space learned by the visual encoder (e.g., CLIP~\citep{radford2021clip} and DINOv2~\citep{oquab_dinov2}) with the language representation space of the LLMs,
different projection modules are proposed and optimized with well-designed training recipes.
%Flamingo~\citep{alayrac2022flamingo} uses the Resampler~\citep{jaegle2021resampler} to extract a fix number of visual tokens from the visual encoder.
%BLIP2~\citep{li2023blip2} first propose the Q-former module to bridge the modality gap between the visual encoder and the LLMs.
Resampler~\citep{jaegle2021resampler} and Q-former~\citep{li2023blip2} use a group of learnable query vectors to extract the textual-aligned visual embedding by the cross-attention mechanism, independent of input image resolution.
Different from these elaborate projector, LLaVA~\citep{liu2023llava, liu2023llava1.5} series only use the linear MLP as the projector and gets better performance.
For the purpose of improving the efficiency of MLLMs, reducing the number of visual tokens is a effectual way and a group of methods proposed.
PruMerge~\citep{shang2024llava_prumerge} directly prunes unimportant visual tokens guided by the attention scores of the visual encoder, and merges these tokens with similarity scores.
$M^3$~\citep{cai2024m3} aims to learn a MLLM which can capture information from mutli-granularity, variable-length visual token sequences, so that the computational cost of it can be controlled.
C-Abstractor~\citep{cha2024honeybee} adopts the Residual Blocks and average pooling layer to preserve the local context from visual features.
LDPv2~\citep{chu2024mobilevlmv2} takes both the point-wise and the depth-wise convolution layers for the purpose of positional information enhancement.
Tokenpacker~\citep{li2024tokenpacker} introduces a region-to-point injection module to inject the fine-grained region features into the point queries, which reduces the number of visual tokens as a result.
Our SAEP projector adopts the convolution layer to simultaneously
serve the function of spatial understanding enhancement and token reduction.

\section{Approach}

\subsection{Preliminary Discussion}
%%% 介绍mllm的基本结构，给出llm的自回归公式，说明为什么token数量会影响计算效率
%%% mllm能够理解并推理输入的图文模态的信息，以自回归的方式生成回复内容
MLLMs are developed to process multimodal inputs following the instructions and generate reasonable responses.
A typically MLLM consist of three comments: a visual encoder $E_v$, a vision-language projector $P$  and a LLM $\Phi$. %illustrated in \textcolor{red}{Figure}.
%The ViT-based models are widely-used as the visual encoder of MLLMs 
Firstly, the visual encoder (e.g. CLIP~\citep{radford2021clip}, DINOv2~\citep{oquab_dinov2}) takes the image $I_{img} \in \mathbb{R}^{3\times H \times W}$ as input and outputs a patch-level visual feature sequence $I_v \in \mathbb{R}^{N \times C}$, where $N$ is the number of visual features and $C$ is the size of each visual feature.
Then, the projector projects the captured visual features $I_v$ into the token embedding space $E_t$ modeled by the LLM. 
So that the LLM can understand and reason the visual information as the same way as the textual sequence $T$, and then generate the response $Y=\{r_i\}_{i=1}^{|Y|}$ in an autoregressive manner.
The above process can be formalized as:
\begin{equation}
\begin{aligned}
    p_{\Phi}(Y | T_v, T_t) = \prod_{i=1}^{|Y|} p_{\Phi}(y_i|,  y_{<i}, T_v, T_t)
\end{aligned}
\end{equation}
where $T_v = P(E_v(I_{img}))$ and $T_t =E_t(T)$.

Due to the dominant parameter of LLM, it is primarily responsible for the majority of the computation and memory usage for MLLM.
Taking the quadratic relationship between the computational expense of LLM and the number of input tokens into consideration,
reducing the number of visual tokens becomes a reasonable and effective for improving the efficiency of MLLM.
In addition, 
the projector is tasked with bridging the modality gap between the two-dimensional spatial visual signal and the one-dimensional sequential textual signal.
Thus, the effectiveness of the projector significantly impacts the performance and efficiency of MLLM.

\subsection{Multi-level based Spatial-aware Efficient Projector}
%%% 2d 重构
%%% 多层特征提供更样信息，更多底层局部细节信息
%%% 改进的可分离卷积： 1）point-wise and depthwise convolution 减少参数； 2）卷积的同时压缩 3）avg pooling做残差连接
%%
%% 扣住压缩token的核心目标，我们的方案是卷积压缩token（现有方案是pooling或者attention），引入多层解决压缩导致的原本token序列中空间信息被破坏的问题

Since the MLP projector could neither control the number of visual tokens nor capture the visual spatial knowledge from the visual features,
we propose a multi-level spatial and locality-enhanced efficient projector, named SAEP, to alleviate these two problems.
%The SAEP projector can not only shorten the visual token sequence but also greatly improve the spatial understanding abilities of MLLMs.
%%Our findings reveal that the shallow layer features of CLIP offer particular advantages for fine-grained tasks such as grounding and region understanding
%shallow representations lack sufficient semantic information
%Our analysis reveals that different layers of features exhibitvarying biases towards local and global patterns. Shallow layer features containing low-level detailed information prove beneficial for fine-grained perception tasks such as grounding and positioning ability, while deep layer features are superior at global understanding.

Prior works~\citep{li2022vitdet, jiang2023clip_dino} have revealed that different layers of ViT-based model exhibit varying bias towards local and global patterns.
The visual features extracted from shallower layers contain low-level local fine-grained information, while the deep layer features are superior at provide global high-level information.
Hence, %as shown in \textcolor{red}{Fig2},
our SAEP projector takes multi-level visual features as its input so that more detailed and positional visual clues can be provided to the LLMs.
Specifically, $K$ layers of visual features, $\{I_v^1, I_v^2, ... I_v^K\}, I_v^k \in \mathbb{R}^{N \times C}$, are selected from the visual encoder,
incurring no additional computational overhead compared to those single-layer projectors.

%%% 重构成2d feature map，相当于引入 spatial inductive bias
To facilitate the spatial structure implicated in these serialized visual features, 
we explicitly introduce the spatial inductive bias by converting each visual feature sequence into a group of 2D patch-grid feature maps.
Formally, the same-layer visual features $I_v^K$ are reorganized by their original relative position in the image, 
$C$ feature maps is obtained, marked as $\hat{I}_v^k \in \mathbb{R}^{H \times W \times C}$, where $H$ and $W$ is the height and width of each feature map, and $H\times W=N$.
So there are $C \cdot K$ feature maps in total, marked as $\hat{I}_v \in \mathbb{R}^{H\times W \times CK}$.

On the basis of these feature maps, 
the convolution operation is applied because of its inherently advantage in preserving the spatial and positional knowledge.
Instead of the standard convolution, the depthwise separable convolution is adopted so that the number of parameters and the computational complexity can be reduced.
In detail, the pointwise convolution is first applied to vertically aggregate the visual features from the same position but at different levels into a more compact one.
Then the depth-wise convolution operation takes in charge of horizontally aggregate the spatial regional visual features into a information enriched one.
It should be noted that the stride of depthwise convolution is set the same as the kernel size  
to effectively reduce redundant visual information and the size of feature maps.
Applying convolution uniformly across the feature map is also able to minimize information loss caused by uneven down-sampling.
Moreover, an average pooling layer is also applied on the top of the pointwise convolution for regional feature compression, and its output is added to the output of depth-wise convolution operation.
In essence,
the average pooling layer can be seen as a simplified variant of depthwise convolution and acts as a residual-like visual feature shortcut to avoid the information loss.
After the above convolution operation, the feature maps are flatten into a shorter visual token sequence.

%However, compared to the one-to-one transformation of linear projector,
%压缩token会破坏序列中的空间信息，我们引入多层特征

\begin{table*}[ht]
\centering
\begin{tabular}{l|c|cccc|c}
\toprule
\multirow{2}{*}{\textbf{Method}}  & \multirow{2}{*}{\textbf{\# Tokens}} & \textbf{MME} & \textbf{MMB} & \textbf{SEED} & \textbf{VSR} & \multirow{2}{*}{\textbf{Avg.$^N$}}\\
& & POS & SR/OL/PR &  SR/IL & test &  \\
\midrule
LLaVA-1.5 & 576 & 148.3 & 20.0 / 45.7 / 25.0 & 51.0 / 59.6 & 51.5 & 46.7 \\ %% 46.7
\midrule
Resampler & 144 & 93.3 & 24.4 / 35.8 / 37.5 & 44.4 / 51.9 & 52.4 & 41.9 \\ %% 41.9
C-Abstractor & 144 & \textbf{123.3} & 20.0 / 45.7 / 37.5 & 48.6 / 57.2 & 53.5 & 46.3 \\ %% 46.3
%LDPv2 & 144 & 55.4 & \textbf{128.3} & 24.4 / 45.7 / 29.2 & 49.3 / 59.2 \\ %% 
LDPv2 & 144 & 121.7 & 20.0 / 46.9 / 25.0 & 50.8 / 60.1 & 52.6 & 45.2\\
Tokenpacker & 144 & 116.7 & 17.8 / 39.5 / 45.8 & 50.5 / 60.9 & 54.4 & 46.8  \\ %% 46.8
%SAEP & \textbf{57.4} & \textbf{131.7} & \textbf{88.4} / \textbf{51.9} / \textbf{45.8} & \textbf{50.7} / 60.1 \\
%SAEP2 & 55.7 & 121.7 & 81.4 / 48.2 / 50.0 & 65.4 / 60.9 \\
SAEP & 144 & 121.7 & \textbf{24.4} / \textbf{46.9} / \textbf{58.3} & \textbf{53.3} / \textbf{60.9} & \textbf{55.7} & \textbf{51.4} \\ %% 50.5
\bottomrule
\end{tabular}
\caption{Results on the multimodal spatial tasks. ``Avg.$^N$'' refers to the normalized average score on the 7 tasks. ``MLP'' refers to the two-layer MLP projector of LLaVA-1.5\protect\footnotemark. The best performance among all the token reduction methods is marked in \textbf{bold}.}
\label{tab:spatial_task}
\end{table*}

\footnotetext{Most results of MLP (LLaVA-1.5) are officially reported, some missed metrics are evaluated by ourselves with the officially released model checkpoint.}

\section{Experiments}
In this section, massive experiments are conducted to demonstrate the effectiveness of the proposed SAEP projector.
The implementation details of the SAEP projector and the training recipe of the MLLMs are firstly described for reproduction. 
For the purpose of evaluation, both the vision-language spatial tasks and the popular general multimodal benchmarks (e.g., VQAv2, MME, MMBench) are selected.
%Moreover, the ablation studies are also presented for in depth analysis.

\subsection{Baselines}
Due to most of existing vision-language projectors are designed and applied in different MLLMs with different training strategies, 
four most popular projectors are selected and equally employed with the training recipe of LLaVA-1.5 for a fair comparison.

\begin{itemize}
    \item \textbf{Resampler} is the most popular token reduction method. It can produce a pre-set fixed number of visual tokens with the cross-attention mechanism.
    \item \textbf{C-Abstractor} aims to get a balance between the efficiency and performance. The convolution and average pooling operations are introduced in it for locality modeling.
    %introduces the convolution and average pooling operations for locality modeling.
    %is proposed to get a balance between locality preservation and visual token reduction. 
    \item \textbf{LDPv2} utilizes the average pooling layer for compressing the number of visual tokens and introduces a PEG module to enhance positional information.
    \item \textbf{Tokenpacker} injects multiple high-resolution, multi-level visual features into a coarse low-resolution one with a region-to-point attention layer. The official released LLaVA-1.5-tokenpacker checkpoint is used for evaluation because it shares the same training recipe with LLaVA-1.5~\footnote{https://github.com/CircleRadon/TokenPacker}.
    %, and achieves comparable performance with higher computation efficiency compared to LLaVA-1.5.
    
\end{itemize}
\subsection{Implementation Details}
The SAEP projector and all the baselines are optimized with the same training recipe as the LLaVA-1.5.
In detail, 
we take CLIP-ViT-L-336px as the visual encoder and Vicuna-7B as the LLM.
Visual features from the 10th, 14th, 18th, 21st and 23rd layers of CLIP are utilized in the SAEP projector.
Two-stage pre-training pipeline is adopted to get the instruction-following MLLMs.
The LCS-558K image-text pair dataset is used for the first alignment stage pre-training and only the parameters of the projector are optimized while all the others are frozen.
The 665K multi-dataset mixed instruction-following data samples are performed for the second instruction-tuning stage.
The projector and LLM are jointly optimized during this stage, while the visual encoder keeps frozen.
The batch size of the two stage training is set as 256 and 96, respectively.
The AdamW optimizer with a Cosine learning rate schedule is adopted and the initial learning rate is set as 1e-3 and 2e-5 for the two stage training.
The models are trained on 8 $\times$ NVIDIA A100 40GB GPUs.

%%数据，bs，gpu，LLM，visual encoder,图像分辨率
%% 对空间任务做一些说明

%% implementation details需要说明的信息 
%% visual encoder设置，llm设置，两阶段训练，两个阶段都有哪些参数被优化；
%% bs, adam，lr schedule, lr, 训练的gpu信息

%% 对提供的结果做一些说明，tokenpacker模型的结果有些

% \subsection{Evaluation Benchmarks}
% %% 分成三组，空间任务，定位任务，通用任务
% Three groups of multimodal evaluation tasks are performed for comprehensively verifying the effectiveness and capabilities of our SAEP projector.
% The first group of tasks aims to evaluate the visual grounding abilities (i.e., locating the object in the given image based on the textual description), including the Refcoco, Refcoco+ and Refcocog datasets.
% Honeybee adopts 6 tasks to assess the spatial understanding capabilities, including the Position (POS) sub-task of MME, Spatial Relationship (SR), Object Location (OL), and Physical Relation (PR) of MMBench, Spatial Relation (SR) and Instance Location (IL) of SEED-Bench.
% In addition to the above 6 tasks, the Visual Spatial Reasoning task is also included in the second group of evaluation tasks.
% The third group of tasks include both the general vision-language tasks (i.e., VQAv2, GQA, Vizwiz and TextVQA) and MLLM benchmarks (i.e., POPE, MME, MMBench, SEED-Image and MMVet), aiming to evaluate the general vision-langauge understanding capabilities.

\subsection{Experimental Results}
Multiple groups of multimodal evaluation tasks are performed for comprehensively verifying the effectiveness and capabilities of our SAEP projector.

\subsubsection{Performances on Spatial Understanding Tasks}
Honeybee adopts 6 tasks to assess the spatial understanding capabilities, including the
Position (POS) sub-task of MME, Spatial Relationship (SR), Object Location (OL), and Physical Relation (PR) of MMBench, Spatial Relation (SR) and Instance Location (IL) of SEED-Bench.
In addition to the above 6 tasks, the Visual Spatial Reasoning (VSR) task is also included in this group of evaluation tasks.

Table~\ref{tab:spatial_task} provides the results of the baseline projectors on these 7 spatial understanding tasks.
Notably, the SAEP projector demonstrates its advantage on visual spatial understanding with a large average score margin compared to previous SoTA method (51.4 vs. 46.8).

Interestingly, it can also be found that compared to the one-to-one transformed MLP projector, the multi-level visual feature based projectors (i.e., Tokenpacker and SAEP) not only reduce 75\% visual tokens but also get better performance.
On the contrary,
the single-level based token reduction methods (i.e., Resampler, C-Abstractor and LDPv2) underform the MLP projector on these spatial tasks.
This phenomenon further verifies the directly compressing the visual token sequence will disrupt the spatial knowledge implicit in the original sequence and introducing multi-level visual features is a cost-effective choices.

\begin{table*}[ht]
\centering
\begin{tabular}{l|c|ccc|c}
\toprule
\multirow{2}{*}{\textbf{Method}}  & \multirow{2}{*}{\textbf{\#Tokens}} & \textbf{Refcoco} & \textbf{Refcoco+} & \textbf{Refcocog} & \multirow{2}{*}{\textbf{Avg.}} \\
& & val / testA / testB & val / testA / testB & val / test & \\
\midrule
LLaVA-1.5 & 576 & 55.2 / 63.1 / 46.4 & 49.7 / 58.7 / 38.1 & 51.4 / 50.7 & 51.7 \\
\midrule
Resampler & 144 & 31.9 / 34.9 / 28.9 & 21.0 / 26.3 / 18.0 & 25.0 / 24.3 & 26.3\\
C-Abstractor & 144 & 60.0 / 64.7 / 55.5 & 50.5 / 58.2 / 44.6 & 53.3 / 54.7 & 55.2  \\
%LDPv2 & 144 & 61.6 / 65.8 / 56.9 & 54.2 / 60.3 / 47.0 & 55.7 / 57.4 & 57.4 \\
LDPv2 & 144 & 61.5 / 66.4 / 57.0 & 54.1 / 62.2 / 46.7 & 55.6 / 56.9 & 57.6 \\
Tokenpacker & 144 & 60.6 / 65.9 / 56.8 & 53.1 / 61.4 / 46.8 & 55.9 / 56.9 & 57.2 \\
SAEP & 144 & \textbf{65.5} / \textbf{67.6} / \textbf{63.4} & \textbf{58.3} / \textbf{63.5} / \textbf{52.9} & \textbf{59.8} / \textbf{60.3} & \textbf{61.4}  \\
\bottomrule

\end{tabular}
\caption{Accuracy results of the SAEP projector and the baselines on the visual grounding tasks. The Accuracy is calculated based the Intersection Over Union (IOU) ratio between the predicted and actual bounding boxes, with an IOU exceeding 0.5 classified as a true positive. }
\label{tab:location_task}
\end{table*}

\begin{table*}[ht]
\centering
\small
\begin{tabular}{l|c|cccc|ccccc}
\toprule
Method  & \#Token &  VQAv2 & GQA & Vizwiz & VQA$^{T}$ & POPE & MME & MMB & SEED & MMVet \\
\midrule
LLLaVA-1.5 & 576 & 78.5 & 62.0 & 50.0  & 58.2 & 85.9 & 1511/297 & 64.3 & 58.6 & 31.1   \\
\midrule
Resampler & 144 & 73.3 & 56.6 & 50.2 & 52.4 & 83.6 & 1348/259 & 63.1 & 58.0 & 26.0 \\
C-Abstractor & 144 & 76.5 & 59.9 & 51.1 & 56.2 & 84.6  & \textbf{1471}/\underline{293} & 65.0 &  62.4 & 29.8 \\ %复现
%LDPv2 & 78.0 & \underline{61.4} & 45.7  & 57.0 & 84.7 & 1422/\textbf{330} & \textbf{66.8} & 64.8 & 29.1\\
LDPv2 & 144 & 77.8 & 60.7 & 50.0 & \underline{57.1} & \underline{84.9} & 1446/267 & \textbf{66.4} & \underline{65.2} & 28.0 \\
TokenPacker & 144 & \underline{78.1} & \underline{61.1} & \underline{52.0}  & \textbf{58.0}  & \textbf{85.9} & 1432/277 & 65.7 & 64.9 & \textbf{32.8} \\ 
SAEP & 144 & \textbf{78.5} & \textbf{62.4} & \textbf{53.0} & \textbf{58.0} & \textbf{85.9} & \underline{1467}/\textbf{316} & \underline{66.3} & \textbf{65.4} & \underline{30.6} \\
\bottomrule

\end{tabular}
\caption{Results on general multimodal tasks. The best performance on each task is marked with \textbf{bold} and the second-best is marked with \underline{underlined}. The MME benchmark includes two sub-tasks: perception and cognition.}
\label{tab:merge}
\end{table*}

\subsubsection{Performances on Visual Location Tasks}
The first group of evaluation tasks aims to assess the visual grounding abilities (i.e., locating the object in the given image based on the textual description). 
Three most popular visual grounding (i.e., Refcoco, Refcoco+ and Refcocog) benchmarks are adopted.

The evaluation results are provided in Table~\ref{tab:location_task}.
It can be seen that our SAEP projector shows significant performance improvement on the three visual grounding benchmarks.
%Our SAEP projector achieves 9.7 absolute accuracy score gain while reducing 75\% visual tokens compared to the standard MLP projector (51.7).
%Moreover, compare to the SoTA token reduction method (LDPv2 with 57.4), the proposed SAEP projector still gets a large performance margin (+4.0).
%Therefore, the superiority of the SAEP projector on the visual location task is proven.
%Compare to the SoTA token reduction method (LDPv2 with 57.4), the proposed SAEP projector still gets a large performance margin (+4.0).
%More than that, compared the MLP projector
Compare to the SoTA token reduction method (LDPv2 with 57.4), the proposed SAEP projector gets a large performance margin (+4.0).
Moreover, multiple token reduction projectors get better performance than the standard MLP projector, which indicates that the projector not only projects the visual feature to the token embedding space, but also plays an critical role in extracting visual semantics for the LLM backbone.
Among them, our SAEP projector performs best and achieves 9.7 absolute accuracy score gain while reducing 75\% visual tokens compared to the standard MLP projector (51.7).

When taking the projector architecture into account,
the effectiveness of the SAEP projector's design on modeling the spatial and locational knowledge can be further verified.
In specific,  
although the LDPv2 is the same convolution-based method as the SAEP projector,
the SAEP projector further benefits form the leveraging of multi-level features.
However, even though the Tokenpacker utilizes the multi-level fine-grained features to enhance the performance,
its cross-attention based architecture makes it disadvantaged at extracting the spatial and locational knowledge compared to the convolution-based SAEP projector.
%Overall, the superiority of the SAEP projector on the visual spatial understanding and location task is proven.

\begin{figure*}[t]
    \centering
    \includegraphics[width=1.0\linewidth]{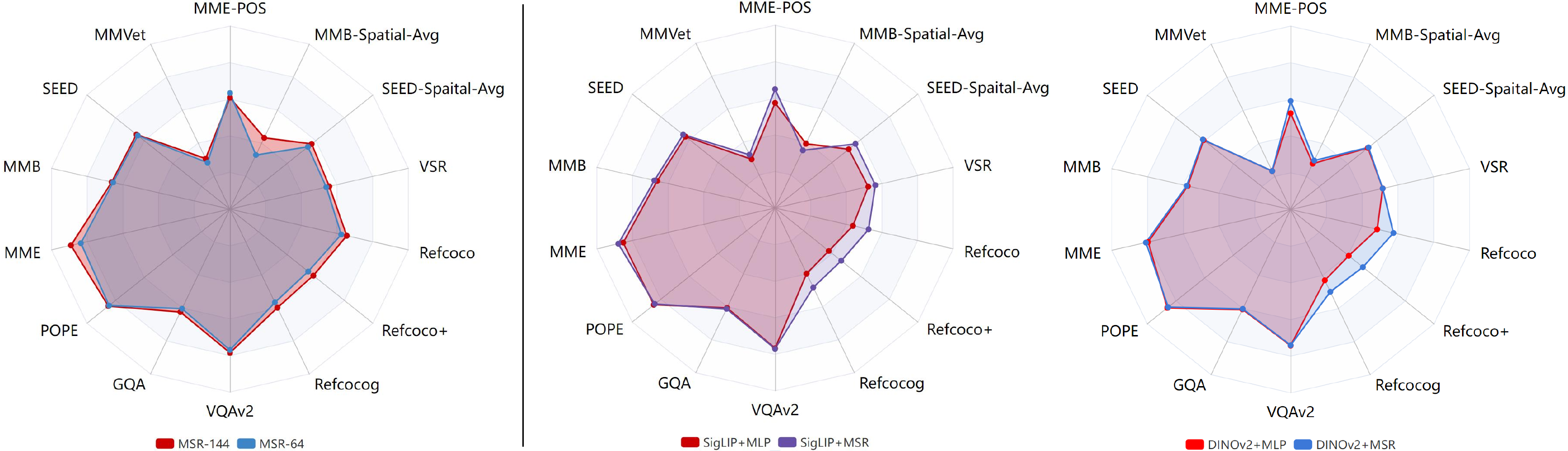}
    (a) ~~~~~~~~~~~~~~~~~~~~~~~~~~~~~~~~~~~~~~~~~~~~~~~~~~~~ ~~~~~~~~~~~~~~~~~~~~~~~~~~~~(b)~~~~~~~~~~~~~~~~~~~~~~~~~~~~~~
    \caption{(a) The performance of the SAEP projector with different number of visual tokens. (b) The performance of the MLP and our SAEP projector with different visual encoders. ``MMB-Sptial-Avg'' and ``SEED-Spaital-Avg'' are the average score of MMB and SEED's spatial sub-tasks.}
    \label{fig: radar}
\end{figure*}

\subsubsection{Performances on General Multimodal Task}
The third group of tasks includes both the general vision-language tasks (i.e., VQAv2, GQA, Vizwiz and TextVQA) and popular instruction-following MLLM benchmarks (i.e., POPE, MME, MMBench, SEED-Image and MMVet), aiming to evaluate the general vision-language understanding capabilities.

The results are shown in Table~\ref{tab:merge}.
Compared to the baselines, the SAEP projector gets best performance on 5 of 10 tasks and gets second-best on another 5 tasks. 
That indicates that the SAEP projector has the ability to preserve sufficient global visual semantic information for the MLLM while significantly reducing the number of visual token.

Overall, the results in Table~\ref{tab:merge} shows that as a token reduction projector,
the computational efficiency brought by the SAEP projector does not come at the cost of the multimodal understanding and reasoning abilities of the MLLM.
More than that, the performance shown in Table~\ref{tab:spatial_task} and Table~\ref{tab:location_task} can further proven the superiority of the SAEP projector on extracting and restructuring visual spatial and locational knowledge.

% \begin{table*}[ht]
% \centering
% \begin{tabular}{l|ccccc}
% \toprule
% Method  &  VQAv2 & GQA & Vizwiz & SQA^I & VQA$^T$  \\
% \midrule
% MLP & 78.5 & 62.0 & 50.0 & 66.8 & 58.2  \\
% \midrule
% Resampler  & 73.3 & 56.6 & 50.2 & 69.8 & 52.4 \\
% C-Abstractor & 76.5 & 59.9 & 51.1 & 70.4 & 56.2 \\ %复现
% LDPv2 & 78.0 & 61.4 & 45.7 & 69.0 & 57.0 \\
% TokenPacker  & 77.9 & 61.1 & 52.0 & \textbf{70.4} & \textbf{58.0} \\ 
% SAEP  & \textbf{78.4} & \textbf{62.0} & \textbf{53.0} & 69.6 & 57.8 \\
% \bottomrule
% \end{tabular}
% \caption{xxx}
% \label{tab:projector1}
% \end{table*}

% \begin{table*}[ht]
% \centering
% \begin{tabular}{l|cccccc}
% \toprule
% Method & POPE & MME & MMB & MMB$^{CN}$ & SEED & MM-Vet \\
% \midrule
% MLP  & 85.9 & 1510.7 / 297.1 & 64.3 & 58.3 & 58.6 & 31.1 \\
% \midrule

% Resampler & 83.6 & 1347.8 / 259.3 & 63.1 & 54.1 & 58.0 & 26.0 \\ %%%我们复现评估
% CAbstractor & 84.6  & \textbf{1471.4} / 292.9 & 65.0 & 56.6 & 62.4 & 29.8 \\ %% 我们复现评估
% LDPv2 &  84.7 & 1422.0 / \textbf{330.0} & \textbf{66.8} & \textbf{59.6} & 64.8 & 29.1  \\
% TokenPacker  & \textbf{85.9} & 1431.6 / 277.1 & 65.7 & 57.9 & \underline{64.9} & \textbf{32.8} \\
% %SAEP & \underline{86.3} & \underline{1448.8} / \textbf{330.7} & \underline{66.1} & \textbf{59.7} & \textbf{65.1} & \underline{31.3} \\
% %SAEP2 & 84.6 & 1430.8 / 277.9 & 66.3 & 58.9 & 65.4 & 30.6 \\
% SAEP & \underline{84.9} & \underline{1466.7} / \underline{321.8} & \underline{66.3} & \underline{58.9} & \textbf{65.4} & \underline{31.3} \\
% \bottomrule
% \end{tabular}
% \caption{xxx}
% \label{tab:projector2}
% \end{table*}

%%
\section{Ablation Study}

\subsection{Efficiency Improvement}
%% pre-train: 3671s; finetune: 28173.9s, 7.8h

\begin{table}[ht]
\centering
\begin{tabular}{l|c|ccc}
\toprule
\textbf{Method} & \textbf{\#Tokens} & \textbf{TPS} & \textbf{PT}  & \textbf{IT} \\
\midrule
MLP & 576 & 29.1 & 3.5h & 10h  \\
SAEP & 144 & 34.2 & 1h & 7.8h \\
SAEP & 64 & 35.2 & 0.6h & 7.4h \\
\bottomrule

\end{tabular}
\caption{Training times with 8 NVIDIA A100 40GB GPUs and same LLaVA training recipe. 
``PT'' and ``IT'' is the abbreviation of ``Pre-training'' and ``Instruction-tuning''.``TPS'' refers to ``token per second''.}
\label{tab:efficiency}
\end{table}

To illustrate the efficiency improvement brought by our SAEP, the comparisons from multiple perspectives between the original MLP projector and our SAEP projector are provided in Table~\ref{tab:efficiency}. 
Moreover, the efficiency of the SAEP projector with 64 visual tokens is also employed and evaluated.
The performance difference are shown in Figure~\ref{fig: radar} (a).
To be clear, the token per second (TPS) is evaluated on the MMVet test set by One A100 GPU.

Table~\ref{tab:efficiency} indicates that our SAEP projector with 144 token can save about 35\% training time compared to the MLP projector, while the SAEP projector with 64 token can further save 40.7\%.
In addition, the results on TPS show that the 144-token and 64-token SAEP projectors boost the inference efficiency by 17.5\% and 21.0\% respectively.
Meanwhile, as illustrated in Figure~\ref{fig: radar} (a),
even though the visual token is reduced by 89\%,
the SAEP projector is still able to perform better than the MLP projector on the spatial and locational tasks, and the performance on the general multimodal tasks is comparable.
Therefore, our SAEP projector can not only significantly improve the computational efficiency, but also enhance the spatial and locational capabilities of the MLLM.

\subsection{Effectiveness on Other Visual Encoder}
\label{sec:other_visual_encoder}
As a token reduction projector, the generalization ability of the proposed SAEP on different visual encoder is very important.
Besides to the adopted CLIP-ViT visual encoder, another two popular visual encoders, SigLIP and DINOv2, are select to verify the SAEP projector's generalization. The training recipe is the same as the LLaVA-1.5 except for the visual encoder.
The input image size of both the SigLIP and DINOv2 is 224px.
%% 雷达图
The results are shown in Figure~\ref{fig: radar}.
It is obvious that with the two different visual encoders,
the efficient SAEP projector still demonstrate its advantage on enhancing the spatial and locational abilities of the MLLM.
It can also be seen that the SAEP projector can get slight performance improvement on the general multimodal tasks.
In a word, the SAEP is an general spatial and locality-enhanced efficient projector.

% \begin{figure}[t]
%     \centering
%     \includegraphics[width=1.0\linewidth]{latex/Siglip-dinov2-SAEP-radar-merge.pdf}

%     \caption{The performance of the MLP and our SAEP projector with different visual encoders. ``MMB-Sptial-Avg'' and ``SEED-Spaital-Avg'' are the average score of MMB and SEED's spatial sub-tasks.}
%     \label{fig: radar}
% \end{figure}

\begin{table*}[ht]
\centering
\begin{tabular}{ccc|ll|ccccc}
\toprule
\textbf{Multi-level} & \textbf{Conv.} & \textbf{Pooling} & S-Avg. & L-Avg. & GQA & VQA$^{T}$ & POPE & MMB & SEED \\
\midrule
$\checkmark$ & $\checkmark$ & $\checkmark$ & \textbf{51.4} & \textbf{61.4} & \textbf{62.4} & \textbf{58.0} & \textbf{85.0} & \textbf{66.3} & \textbf{65.4}  \\
$\checkmark$ & $\times$ & $\checkmark$ & 47.9 \small{(-3.5)} & 61.4 & 61.9 & 57.6 & 84.6 & 66.2 & 65.3 \\
$\checkmark$ & $\checkmark$ & $\times$ &  46.8 \small{(-4.6)} & 58.7 \small{(-2.7)} & 61.1 & 57.8 & 84.9 & 66.0 & 64.9\\
$\times$ & $\checkmark$ & $\checkmark$ & 44.7 \small{(-6.7)} & 57.6 \small{(-3.8)} & 61.2 & 56.8 & 64.8 & 65.6 & 64.6 \\
\bottomrule
\end{tabular}
\caption{Component-wise ablation results of the proposed SAEP projector. ``S-Avg.'' and ``L-Avg.'' refers to the average score of the above adopted visual spatial and location tasks respectively. }
\label{tab:arch_ablation}
\end{table*}

\subsection{Architecture Design Ablation}
There are three main components adopted in the SAEP projector: point-wise convolution with multi-level visual features, a depth-wise convolution layer and an average pooling layer. 
We conduct a group of ablation studies to verify the effectiveness of the SAEP's design and the results are provided in Table~\ref{tab:arch_ablation}.

It can be seen that removing each component will result in a decrease in performance,  which indicate that all the three components are helpful to the final performance improvement.
Besides, it can be also found that introducing multi-level visual features makes the greatest contribution to the final performance.
That denotes that providing supplemental detailed local information is beneficial to improve the visual understanding and reasoning abilities, especially for the token reduction projector.
%% 这表明对于压缩token的方案来说，多层特征能有效补充压缩token导致的信息损失，这为后续的projector提供了有价值指导
Meanwhile, the results also reveal that the ``residual'' connection between the depth-wise convolution operation and average pooling operation mainly contributes to capture the spatial and locational information.

%%% 1.移除多层 2.移除avg pooling; 3.换用avg pooling

\begin{figure}[t]
    \centering
    \includegraphics[width=1.0\linewidth]{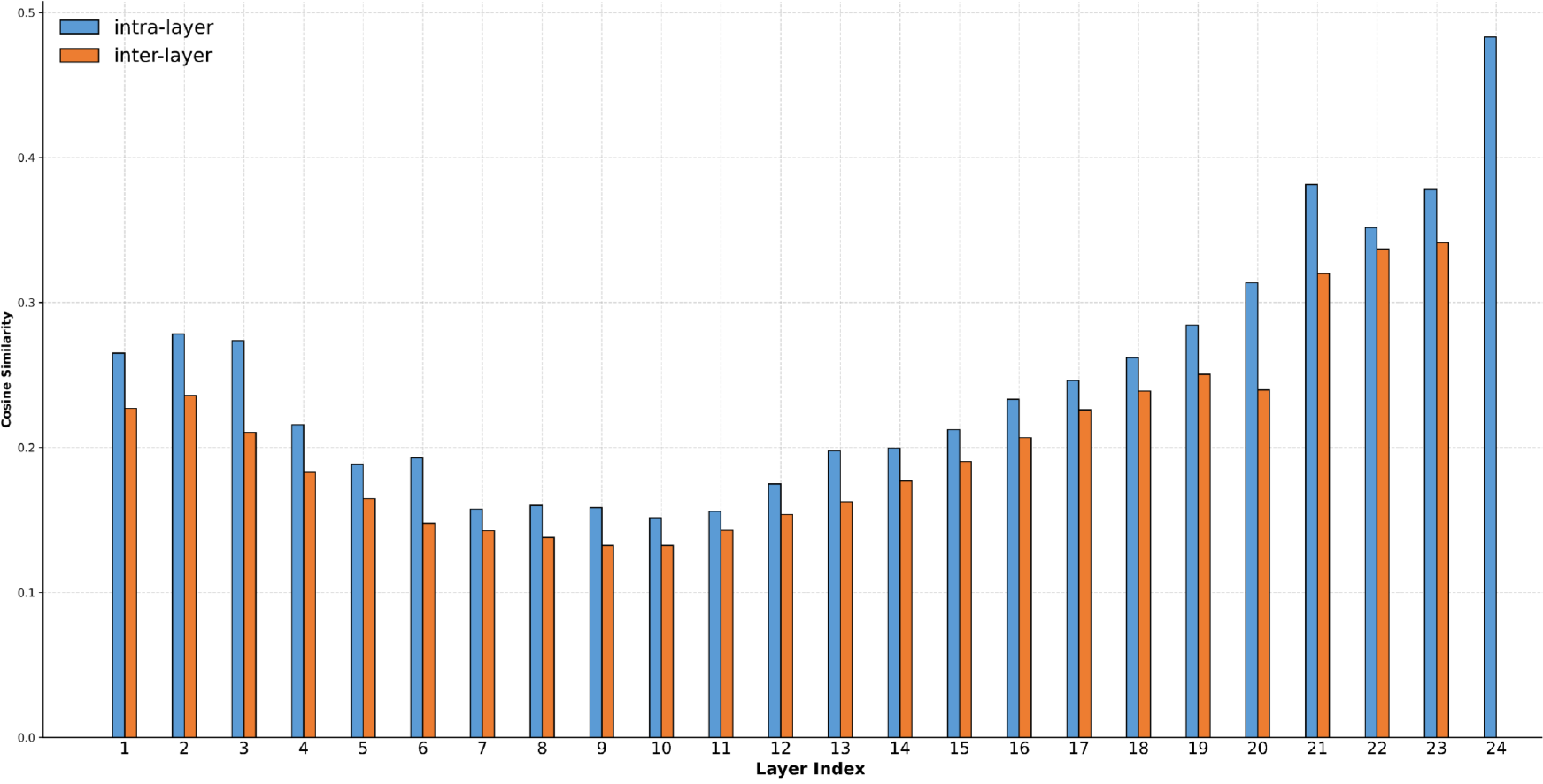}
    \caption{Average cosine similarity over 300 images calculated by the CLIP intra-layer and inter-layer. Inter-layer means the layer $X$ and its adjacent layer $X+1$.}
    \label{fig: layer_selection}
\end{figure}

\begin{figure}[t]
    \centering
    \includegraphics[width=1.0\linewidth]{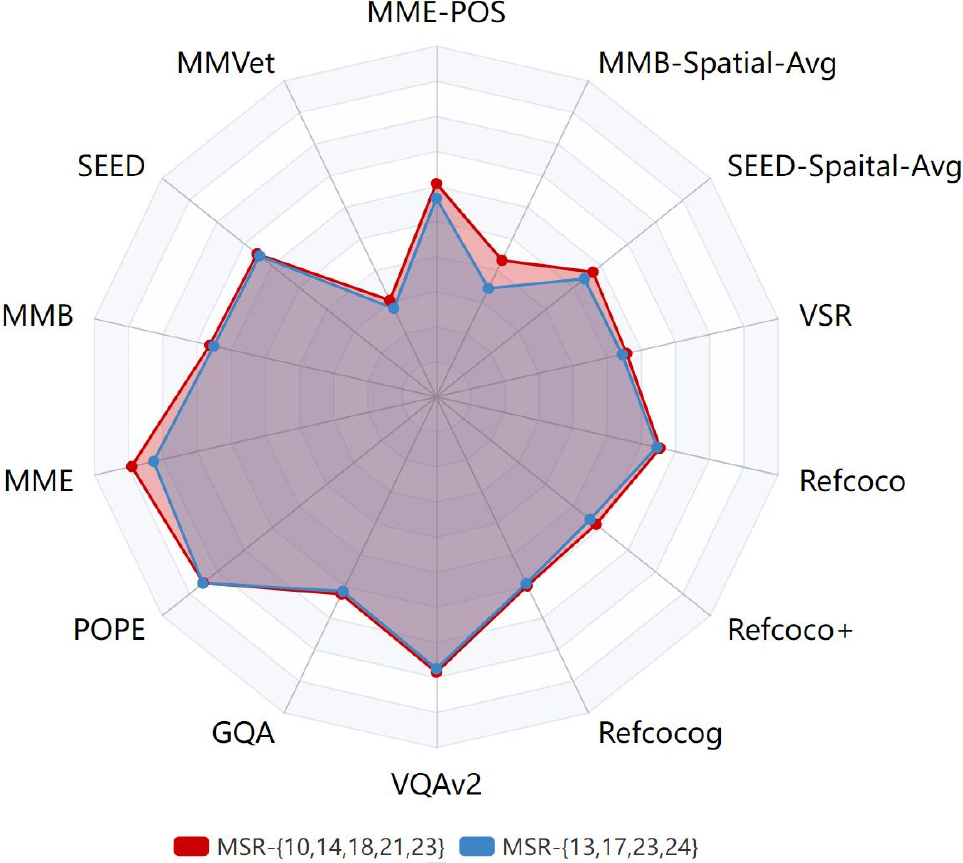}
    \caption{Average cosine similarity over 300 images calculated by the CLIP intra-layer and inter-layer. Inter-layer means the layer $X$ and its adjacent layer $X+1$.}
    \label{fig: layer_comparison}
\end{figure}

\subsection{Exploration on the Layer Selection Strategy}

The SAEP projector utilizes the multi-level visual features to supplementary fine-grained local information.
Even though some recent MLLM works (e.g. TokenPacker) have realized that introducing multi-level visual features is beneficial for the performance.
there is no an available explicit guideline to select the most suitable visual features from different layers.
%about the guidelines of visual encoder layer selection.
%layer selection guidelines.
All these works have to select the features from different layers by experience and iterative testing.

%%% 不同层之间的visual features计算余弦相似度
%Some efforts are made to alleviate this problem.
%We describe the detailed process of the layer selection and hope it is helpful for the future works.
We propose a novel layer selection strategy and verify its effectiveness by a fair comparison.
We firstly estimate the cosine similarity of inter- and intra-layer visual features of CLIP on 100 images.
The average similarity scores of different layers are visualized in Figure~\ref{fig: layer_selection}.
The most obvious characteristic is that as the number of layers increases, the intra-layer and inter-layer similarity both first decreases and then increases.
The forward propagation of CLIP is a process where the visual information gradually becomes more concrete and then more abstract.
In other words, as the number of layers increases, the model first gets a accurate understanding of each path-level local visual feature.
Then, with the layer number increases, the model begins to aggregate the local detailed information and capture more abstract, global information.
This is also consistent with the viewpoint of recent research~\citep{jiang2023from_clip}.
Therefore, the 10th layer is selected because its intra-layer similarity is the lowest.

Meanwhile, an anomalous layer can be observed in Figure~\ref{fig: layer_selection}.
The 21sh layer exhibits high intra-layer similarity, low similarity with the 20th layer, but high similarity with the 22nd layer. 
This indicates that this layer is the information pivot and it captures the most generalized global information, the visual features captured by the deeper layer are more optimization-specific.
Thus, the 21st layer is selected.
Additionally, our evaluation shows that the similarity decreases
as the layer distance increases.
Consequently, we uniformly sample the 14th and 18th layers between the 10th and 21st layers, and include the commonly used 23rd layer, forming the set of all selected layers.

To verify the effectiveness of the above layer selection strategy,
we make a fair comparison between the set of \{10,14,18,21,23\} layers with the set of \{13,17,23,24\} layers, the former layer set is adopted by the TokenPacker and gets best performance among multiple layer sets.
The comparison results are shown in Figure~\ref{fig: layer_comparison}, our layer selection method get better performance.
In fact, the ablation experiment conducted in Section~\ref{sec:other_visual_encoder}, which applies the SAEP projector on the DINOv2 and SigLIP model, is also guided by the above layer selection strategy and obtains significant performance improvement as shown in Figure~\ref{sec:other_visual_encoder}.
%employs the DINOv2 as the visual encoder and applies the SAEP projector on it,
These results not only demonstrate the performance gain brought by an optimal layer sets, but also highlight the worth of our layer selection strategy.

\section{Conclusion}
%% 尽管projector在mllm中承担的至关重要的作用，对其深入的分析和探索还比较欠缺；
%% 为了解决目前已有方案忽略图像序列和文本序列空间信息不一致的问题，我们提出了一种基于多层视觉特征的spatial-aware efficient projector (SAEP).
%% 实验结果表明，我们的SAEP不但能显著降低visual token数量，显著提高mllm的训练和推理效率，还能显著增强mllm的空间信息理解能力。此外，在目前最流行的多模态评估benchmark上，我们的SAEP也表现出了强大的性能。
%% 深入的分析表明不仅表明SAEP具备对其他visual encoder的强大兼容能力和也证明了我们结构设计的有效性。

%% multi-layer visual features对改善mllm表现有巨大帮助，
%%我们提出了一种层特征选择策略，希望能对后续的研究工作指明方向。

%% 我们的SAEP具备对其他visual encoder的强大兼容能力和。
%% SAEP

Despite the crucial role played by the projector in multi-modal language models (MLLMs), its in-depth analysis and exploration remain limited. To address the issue of spatial information discrepancy between 2D visual token sequences and textual token sequences, we propose a spatially aware efficient projector (SAEP) based on multi-layer visual features.
Experimental results demonstrate that our SAEP not only significantly reduces the number of visual tokens, thereby substantially improving the training and inference efficiency of MLLMs, but also significantly enhances the MLLM's ability to understand spatial information. Furthermore, our SAEP exhibits strong performance on the most popular multi-modal evaluation benchmarks compared to existing projectors.
Further analysis not only highlights the strong compatibility of SAEP with other visual encoders but also validates the effectiveness of our architectural design.

%%%%%%%%%%%%%%%%%%%%%%%%%%%%%%%%%%%%%%%%%%%%%%%%%%%%%%%%%%%%%%%%%%%%%%%%%%%%%%%%%%%%%%%%%%%%%%%%%%%%%%%%%%%%%%%%%%%%%%%%%%%%%%%%%%%%%%%%%%%%%%%%%%%%%%%%%%%%%%%%%%%%%%%%%%%%%%%%%%%%%%%%%%%%%%%%%%%%%%%%%%%%%%%%%%%%%%%%%%%%%%%%%%%%%%%%%%%%%%%%%%%%%%%%%%%%%%%%%%%%%%%%%%%%%%%%%%%%%%%%%%%%%%%%%%%%%%%%%%%%%%%%%%%%%%%%%%%%%%%%%%%%%%%%%%%%%%%%%%%%%%%%%%%%%%%%%%%%%%%%%%%%%%%%%%%%%%%%%%%%%%%%%%%%%%%%%%%%%%%%%%%

% Bibliography entries for the entire Anthology, followed by custom entries
%\bibliography{anthology,custom}
% Custom bibliography entries only
\bibliography{custom}

\begin{thebibliography}{36}
\providecommand{\natexlab}[1]{#1}

\bibitem[{Achiam et~al.(2023)Achiam, Adler, Agarwal, Ahmad, Akkaya, Aleman, Almeida, Altenschmidt, Altman, Anadkat et~al.}]{achiam2023gpt}
Josh Achiam, Steven Adler, Sandhini Agarwal, Lama Ahmad, Ilge Akkaya, Florencia~Leoni Aleman, Diogo Almeida, Janko Altenschmidt, Sam Altman, Shyamal Anadkat, et~al. 2023.
\newblock Gpt-4 technical report.
\newblock \emph{arXiv preprint arXiv:2303.08774}.

\bibitem[{Alayrac et~al.(2022)Alayrac, Donahue, Luc, Miech, Barr, Hasson, Lenc, Mensch, Millican, Reynolds et~al.}]{alayrac2022flamingo}
Jean-Baptiste Alayrac, Jeff Donahue, Pauline Luc, Antoine Miech, Iain Barr, Yana Hasson, Karel Lenc, Arthur Mensch, Katherine Millican, Malcolm Reynolds, et~al. 2022.
\newblock Flamingo: a visual language model for few-shot learning.
\newblock \emph{Advances in neural information processing systems}, 35:23716--23736.

\bibitem[{Bai et~al.(2024)Bai, Geng, Mangalam, Bar, Yuille, Darrell, Malik, and Efros}]{bai2024sequential}
Yutong Bai, Xinyang Geng, Karttikeya Mangalam, Amir Bar, Alan~L Yuille, Trevor Darrell, Jitendra Malik, and Alexei~A Efros. 2024.
\newblock Sequential modeling enables scalable learning for large vision models.
\newblock In \emph{Proceedings of the IEEE/CVF Conference on Computer Vision and Pattern Recognition}, pages 22861--22872.

\bibitem[{Cai et~al.(2024)Cai, Yang, Gao, and Lee}]{cai2024m3}
Mu~Cai, Jianwei Yang, Jianfeng Gao, and Yong~Jae Lee. 2024.
\newblock Matryoshka multimodal models.
\newblock \emph{arXiv preprint arXiv:2405.17430}.

\bibitem[{Cha et~al.(2024)Cha, Kang, Mun, and Roh}]{cha2024honeybee}
Junbum Cha, Wooyoung Kang, Jonghwan Mun, and Byungseok Roh. 2024.
\newblock Honeybee: Locality-enhanced projector for multimodal llm.
\newblock In \emph{Proceedings of the IEEE/CVF Conference on Computer Vision and Pattern Recognition}, pages 13817--13827.

\bibitem[{Chen et~al.(2022)Chen, Wang, Changpinyo, Piergiovanni, Padlewski, Salz, Goodman, Grycner, Mustafa, Beyer et~al.}]{chenpali}
Xi~Chen, Xiao Wang, Soravit Changpinyo, AJ~Piergiovanni, Piotr Padlewski, Daniel Salz, Sebastian Goodman, Adam Grycner, Basil Mustafa, Lucas Beyer, et~al. 2022.
\newblock Pali: A jointly-scaled multilingual language-image model.
\newblock In \emph{The Eleventh International Conference on Learning Representations}.

\bibitem[{Chiang et~al.(2023)Chiang, Li, Lin, Sheng, Wu, Zhang, Zheng, Zhuang, Zhuang, Gonzalez, Stoica, and Xing}]{vicuna2023}
Wei-Lin Chiang, Zhuohan Li, Zi~Lin, Ying Sheng, Zhanghao Wu, Hao Zhang, Lianmin Zheng, Siyuan Zhuang, Yonghao Zhuang, Joseph~E. Gonzalez, Ion Stoica, and Eric~P. Xing. 2023.
\newblock \href {https://lmsys.org/blog/2023-03-30-vicuna/} {Vicuna: An open-source chatbot impressing gpt-4 with 90\%* chatgpt quality}.

\bibitem[{Chu et~al.(2024{\natexlab{a}})Chu, Qiao, Zhang, Xu, Wei, Yang, Sun, Hu, Lin, Zhang et~al.}]{chu2024ldpv2}
Xiangxiang Chu, Limeng Qiao, Xinyu Zhang, Shuang Xu, Fei Wei, Yang Yang, Xiaofei Sun, Yiming Hu, Xinyang Lin, Bo~Zhang, et~al. 2024{\natexlab{a}}.
\newblock Mobilevlm v2: Faster and stronger baseline for vision language model.
\newblock \emph{arXiv preprint arXiv:2402.03766}.

\bibitem[{Chu et~al.(2024{\natexlab{b}})Chu, Qiao, Zhang, Xu, Wei, Yang, Sun, Hu, Lin, Zhang et~al.}]{chu2024mobilevlmv2}
Xiangxiang Chu, Limeng Qiao, Xinyu Zhang, Shuang Xu, Fei Wei, Yang Yang, Xiaofei Sun, Yiming Hu, Xinyang Lin, Bo~Zhang, et~al. 2024{\natexlab{b}}.
\newblock Mobilevlm v2: Faster and stronger baseline for vision language model.
\newblock \emph{arXiv preprint arXiv:2402.03766}.

\bibitem[{Dong et~al.(2024{\natexlab{a}})Dong, Zhang, Zang, Cao, Wang, Ouyang, Wei, Zhang, Duan, Cao et~al.}]{dong2024internlm-xcomposer2}
Xiaoyi Dong, Pan Zhang, Yuhang Zang, Yuhang Cao, Bin Wang, Linke Ouyang, Xilin Wei, Songyang Zhang, Haodong Duan, Maosong Cao, et~al. 2024{\natexlab{a}}.
\newblock Internlm-xcomposer2: Mastering free-form text-image composition and comprehension in vision-language large model.
\newblock \emph{arXiv preprint arXiv:2401.16420}.

\bibitem[{Dong et~al.(2024{\natexlab{b}})Dong, Zhang, Zang, Cao, Wang, Ouyang, Zhang, Duan, Zhang, Li et~al.}]{dong2024internlm2-4khd}
Xiaoyi Dong, Pan Zhang, Yuhang Zang, Yuhang Cao, Bin Wang, Linke Ouyang, Songyang Zhang, Haodong Duan, Wenwei Zhang, Yining Li, et~al. 2024{\natexlab{b}}.
\newblock Internlm-xcomposer2-4khd: A pioneering large vision-language model handling resolutions from 336 pixels to 4k hd.
\newblock \emph{arXiv preprint arXiv:2404.06512}.

\bibitem[{Dosovitskiy(2020)}]{dosovitskiy2020vit}
Alexey Dosovitskiy. 2020.
\newblock An image is worth 16x16 words: Transformers for image recognition at scale.
\newblock \emph{arXiv preprint arXiv:2010.11929}.

\bibitem[{Gao et~al.(2024)Gao, Zhang, Liu, Qiu, Huang, Lin, Zhao, Geng, Lin, Jin et~al.}]{gao2024sphinx-x}
Peng Gao, Renrui Zhang, Chris Liu, Longtian Qiu, Siyuan Huang, Weifeng Lin, Shitian Zhao, Shijie Geng, Ziyi Lin, Peng Jin, et~al. 2024.
\newblock Sphinx-x: Scaling data and parameters for a family of multi-modal large language models.
\newblock \emph{CoRR}.

\bibitem[{Jaegle et~al.(2021)Jaegle, Gimeno, Brock, Vinyals, Zisserman, and Carreira}]{jaegle2021resampler}
Andrew Jaegle, Felix Gimeno, Andy Brock, Oriol Vinyals, Andrew Zisserman, and Joao Carreira. 2021.
\newblock Perceiver: General perception with iterative attention.
\newblock In \emph{International conference on machine learning}, pages 4651--4664. PMLR.

\bibitem[{Jiang et~al.(2024)Jiang, Sablayrolles, Roux, Mensch, Savary, Bamford, Chaplot, Casas, Hanna, Bressand et~al.}]{jiang2024mixtral}
Albert~Q Jiang, Alexandre Sablayrolles, Antoine Roux, Arthur Mensch, Blanche Savary, Chris Bamford, Devendra~Singh Chaplot, Diego de~las Casas, Emma~Bou Hanna, Florian Bressand, et~al. 2024.
\newblock Mixtral of experts.
\newblock \emph{arXiv preprint arXiv:2401.04088}.

\bibitem[{Jiang et~al.(2023{\natexlab{a}})Jiang, Liu, Liu, Zhang, Li, Xiong, and Tian}]{jiang2023clip_dino}
Dongsheng Jiang, Yuchen Liu, Songlin Liu, Xiaopeng Zhang, Jin Li, Hongkai Xiong, and Qi~Tian. 2023{\natexlab{a}}.
\newblock From clip to dino: Visual encoders shout in multi-modal large language models.

\bibitem[{Jiang et~al.(2023{\natexlab{b}})Jiang, Liu, Liu, Zhang, Li, Xiong, and Tian}]{jiang2023from_clip}
Dongsheng Jiang, Yuchen Liu, Songlin Liu, Xiaopeng Zhang, Jin Li, Hongkai Xiong, and Qi~Tian. 2023{\natexlab{b}}.
\newblock From clip to dino: Visual encoders shout in multi-modal large language models.

\bibitem[{Karamcheti et~al.(2024)Karamcheti, Nair, Balakrishna, Liang, Kollar, and Sadigh}]{karamcheti2024prismatic}
Siddharth Karamcheti, Suraj Nair, Ashwin Balakrishna, Percy Liang, Thomas Kollar, and Dorsa Sadigh. 2024.
\newblock Prismatic vlms: Investigating the design space of visually-conditioned language models.
\newblock In \emph{Forty-first International Conference on Machine Learning}.

\bibitem[{Li et~al.(2023{\natexlab{a}})Li, Zhang, Yang, Zhang, Pu, and Liu}]{li2023otterhd}
Bo~Li, Peiyuan Zhang, Jingkang Yang, Yuanhan Zhang, Fanyi Pu, and Ziwei Liu. 2023{\natexlab{a}}.
\newblock Otterhd: A high-resolution multi-modality model.
\newblock \emph{arXiv preprint arXiv:2311.04219}.

\bibitem[{Li et~al.(2024{\natexlab{a}})Li, Zhang, Zhang, Zhang, Li, Li, Ma, and Li}]{li2024llava-next-interleave}
Feng Li, Renrui Zhang, Hao Zhang, Yuanhan Zhang, Bo~Li, Wei Li, Zejun Ma, and Chunyuan Li. 2024{\natexlab{a}}.
\newblock Llava-next-interleave: Tackling multi-image, video, and 3d in large multimodal models.
\newblock \emph{arXiv preprint arXiv:2407.07895}.

\bibitem[{Li et~al.(2023{\natexlab{b}})Li, Li, Savarese, and Hoi}]{li2023blip2}
Junnan Li, Dongxu Li, Silvio Savarese, and Steven Hoi. 2023{\natexlab{b}}.
\newblock Blip-2: Bootstrapping language-image pre-training with frozen image encoders and large language models.
\newblock \emph{arXiv preprint arXiv:2301.12597}.

\bibitem[{Li et~al.(2024{\natexlab{b}})Li, Yuan, Liu, Tang, Wang, Zhu, and Zhang}]{li2024tokenpacker}
Wentong Li, Yuqian Yuan, Jian Liu, Dongqi Tang, Song Wang, Jianke Zhu, and Lei Zhang. 2024{\natexlab{b}}.
\newblock Tokenpacker: Efficient visual projector for multimodal llm.
\newblock \emph{arXiv preprint arXiv:2407.02392}.

\bibitem[{Li et~al.(2022)Li, Mao, Girshick, and He}]{li2022vitdet}
Yanghao Li, Hanzi Mao, Ross Girshick, and Kaiming He. 2022.
\newblock Exploring plain vision transformer backbones for object detection.
\newblock In \emph{European conference on computer vision}, pages 280--296. Springer.

\bibitem[{Li et~al.(2024{\natexlab{c}})Li, Zhang, Wang, Zhong, Chen, Chu, Liu, and Jia}]{li2024mini-genimi}
Yanwei Li, Yuechen Zhang, Chengyao Wang, Zhisheng Zhong, Yixin Chen, Ruihang Chu, Shaoteng Liu, and Jiaya Jia. 2024{\natexlab{c}}.
\newblock Mini-gemini: Mining the potential of multi-modality vision language models.
\newblock \emph{arXiv preprint arXiv:2403.18814}.

\bibitem[{Li et~al.(2024{\natexlab{d}})Li, Yang, Liu, Ma, Zhang, Yang, Sun, Liu, and Bai}]{li2024monkey}
Zhang Li, Biao Yang, Qiang Liu, Zhiyin Ma, Shuo Zhang, Jingxu Yang, Yabo Sun, Yuliang Liu, and Xiang Bai. 2024{\natexlab{d}}.
\newblock Monkey: Image resolution and text label are important things for large multi-modal models.
\newblock In \emph{Proceedings of the IEEE/CVF Conference on Computer Vision and Pattern Recognition}, pages 26763--26773.

\bibitem[{Liu et~al.(2023{\natexlab{a}})Liu, Li, Li, and Lee}]{liu2023llava1.5}
Haotian Liu, Chunyuan Li, Yuheng Li, and Yong~Jae Lee. 2023{\natexlab{a}}.
\newblock Improved baselines with visual instruction tuning.

\bibitem[{Liu et~al.(2024)Liu, Li, Li, Li, Zhang, Shen, and Lee}]{liu2024llava-next}
Haotian Liu, Chunyuan Li, Yuheng Li, Bo~Li, Yuanhan Zhang, Sheng Shen, and Yong~Jae Lee. 2024.
\newblock Llava-next: Improved reasoning, ocr, and world knowledge.

\bibitem[{Liu et~al.(2023{\natexlab{b}})Liu, Li, Wu, and Lee}]{liu2023llava}
Haotian Liu, Chunyuan Li, Qingyang Wu, and Yong~Jae Lee. 2023{\natexlab{b}}.
\newblock Visual instruction tuning.
\newblock In \emph{NeurIPS}.

\bibitem[{Oquab et~al.()Oquab, Darcet, Moutakanni, Vo, Szafraniec, Khalidov, Fernandez, HAZIZA, Massa, El-Nouby et~al.}]{oquab_dinov2}
Maxime Oquab, Timoth{\'e}e Darcet, Th{\'e}o Moutakanni, Huy~V Vo, Marc Szafraniec, Vasil Khalidov, Pierre Fernandez, Daniel HAZIZA, Francisco Massa, Alaaeldin El-Nouby, et~al.
\newblock Dinov2: Learning robust visual features without supervision.
\newblock \emph{Transactions on Machine Learning Research}.

\bibitem[{Radford et~al.(2021)Radford, Kim, Hallacy, Ramesh, Goh, Agarwal, Sastry, Askell, Mishkin, Clark et~al.}]{radford2021clip}
Alec Radford, Jong~Wook Kim, Chris Hallacy, Aditya Ramesh, Gabriel Goh, Sandhini Agarwal, Girish Sastry, Amanda Askell, Pamela Mishkin, Jack Clark, et~al. 2021.
\newblock Learning transferable visual models from natural language supervision.
\newblock In \emph{International conference on machine learning}, pages 8748--8763. PMLR.

\bibitem[{Shang et~al.(2024)Shang, Cai, Xu, Lee, and Yan}]{shang2024llava_prumerge}
Yuzhang Shang, Mu~Cai, Bingxin Xu, Yong~Jae Lee, and Yan Yan. 2024.
\newblock Llava-prumerge: Adaptive token reduction for efficient large multimodal models.
\newblock \emph{arXiv preprint arXiv:2403.15388}.

\bibitem[{Team(2024)}]{team2024chameleon}
Chameleon Team. 2024.
\newblock Chameleon: Mixed-modal early-fusion foundation models.
\newblock \emph{arXiv preprint arXiv:2405.09818}.

\bibitem[{Touvron et~al.(2023)Touvron, Martin, Stone, Albert, Almahairi, Babaei, Bashlykov, Batra, Bhargava, Bhosale et~al.}]{touvron2023llama2}
Hugo Touvron, Louis Martin, Kevin Stone, Peter Albert, Amjad Almahairi, Yasmine Babaei, Nikolay Bashlykov, Soumya Batra, Prajjwal Bhargava, Shruti Bhosale, et~al. 2023.
\newblock Llama 2: Open foundation and fine-tuned chat models.
\newblock \emph{arXiv preprint arXiv:2307.09288}.

\bibitem[{Yang et~al.(2024)Yang, Yang, Hui, Zheng, Yu, Zhou, Li, Li, Liu, Huang et~al.}]{yang2024qwen2}
An~Yang, Baosong Yang, Binyuan Hui, Bo~Zheng, Bowen Yu, Chang Zhou, Chengpeng Li, Chengyuan Li, Dayiheng Liu, Fei Huang, et~al. 2024.
\newblock Qwen2 technical report.
\newblock \emph{arXiv preprint arXiv:2407.10671}.

\bibitem[{Zhai et~al.(2023)Zhai, Mustafa, Kolesnikov, and Beyer}]{zhai2023siglip}
Xiaohua Zhai, Basil Mustafa, Alexander Kolesnikov, and Lucas Beyer. 2023.
\newblock Sigmoid loss for language image pre-training.
\newblock In \emph{Proceedings of the IEEE/CVF International Conference on Computer Vision}, pages 11975--11986.

\bibitem[{Zhang et~al.(2023)Zhang, Wang, Cao, Xu, Ouyang, Zhao, Ding, Zhang, Duan, Yan et~al.}]{zhang2023internlm-xcomposer}
Pan Zhang, Xiaoyi Dong~Bin Wang, Yuhang Cao, Chao Xu, Linke Ouyang, Zhiyuan Zhao, Shuangrui Ding, Songyang Zhang, Haodong Duan, Hang Yan, et~al. 2023.
\newblock Internlm-xcomposer: A vision-language large model for advanced text-image comprehension and composition.
\newblock \emph{arXiv preprint arXiv:2309.15112}.

\end{thebibliography}

\end{document}